\documentclass[conference]{IEEEtran}
\IEEEoverridecommandlockouts
\usepackage{cite}
\usepackage{amsmath,amssymb,amsfonts}
\usepackage{algorithmic}
\usepackage{graphicx}
\usepackage{textcomp}
\usepackage{xcolor}
\usepackage{enumitem}
\usepackage[ruled,vlined,linesnumbered]{algorithm2e}
\newcommand{\etal}{\textit{et al.}}
\def\BibTeX{{\rm B\kern-.05em{\sc i\kern-.025em b}\kern-.08em
    T\kern-.1667em\lower.7ex\hbox{E}\kern-.125emX}}
\newcommand{\blue}[1]{\textcolor{black}{#1}}

\begin{document}

\title{Transparent Contribution Evaluation for Secure Federated Learning on Blockchain}


\author{\IEEEauthorblockN{Shuaicheng Ma}
\IEEEauthorblockA{
\textit{Emory University}\\
Atlanta, US \\
sma30@emory.edu}
\and
\IEEEauthorblockN{Yang Cao}
\IEEEauthorblockA{
\textit{Kyoto University}\\
Kyoto, Japan \\
yang@i.kyoto-u.ac.jp}
\and
\IEEEauthorblockN{Li Xiong}
\IEEEauthorblockA{
\textit{Emory University}\\
Atlanta, US \\
lxiong@emory.edu}
\thanks{Corresponding author: Yang Cao.}
}

\maketitle

\begin{abstract}
Federated Learning is a promising machine learning paradigm when multiple parties collaborate to build a high-quality machine learning model.
Nonetheless, these parties are only willing to participate when given enough incentives, such as a fair reward based on their contributions.
Many studies explored Shapley value based methods to evaluate each party's contribution to the learned model.
However, they commonly assume a semi-trusted server to train the model and evaluate the data owners' model contributions, which lacks transparency and may hinder the success of federated learning in practice.
In this work, we propose a blockchain-based federated learning framework and a protocol to transparently evaluate each participant's contribution.
Our framework protects all parties' privacy in the model building phase and transparently evaluates contributions based on the model updates.
The experiment with the handwritten digits dataset demonstrates that the proposed method can effectively evaluate the contributions. 
\end{abstract}

\begin{IEEEkeywords}
Blockchain, Federated Learning, Contribution Evaluation, Transparency, Privacy
\end{IEEEkeywords}

\section{Introduction}
Federated learning (FL) \cite{mcmahan2017communication} is a promising technology that allows \blue{multiple data owners} to jointly train a model without sharing the raw data.
One of the challenges in federated learning is the incentive problem.
Data owners may not be willing to \blue{participate without proper incentives, e.g., a fair reward based on their contributions to the model.}
Recently, Shapley value (SV)-based contribution evaluation protocols \cite{ghorbani2019data,pmlr-v89-jia19a,9006327, wang2020principled} are proposed to help fairly allocate the profit among the participants in federated learning.

However, the existing studies on contribution evaluation in federated learning commonly assume a semi-trusted server to train the model and evaluate the data owners' model contributions. 
\blue{For the cross-silo federated learning that we consider, the organizations (e.g., banks) are mutually untrusted and potentially competing with each other}. It is challenging to assume such a semi-trusted server.
The lack of transparency in contribution evaluation may prevent data owners' collaboration and hinder the success of federated learning in practice.

A \blue{potential} solution to solve the transparency issue is to resort to blockchain techniques. 
One may put all intermediate training results on a blockchain and implement a smart contract to evaluate the contribution.
However, this solution introduces two new challenges: 1) There is a privacy issue if all blockchain data is public to all participants. 
It has been proven that the intermediate data (e.g., weight updates) in federated learning can leak data owners' private information \cite{NEURIPS2019_60a6c400}. 2) \blue{To address the above privacy issue, we can} adopt the well-studied privacy-preserving federated learning, such as secure aggregation (which masks updates) \blue{\cite{10.1145/3133956.3133982}. However, in that case}, the existing Shapley Value based contribution evaluation methods are not compatible since they cannot work on the masked updates (which determine the contributions).

We make the following contributions in this work to address the above challenges. 
\begin{itemize}[noitemsep,topsep=0pt]
    \item We design a blockchain-based secure federated learning framework that adopts secure aggregation to protect data owners' privacy during the training.
    \item We propose a new group-based SV \blue{computation framework} that is compatible with secure aggregation. 
    \item \blue{We implement the framework on the blockchain, and our experiments on real datasets demonstrate the proposed method can effectively evaluate the contributions. }
\end{itemize}

\section{Related Work}
\subsection{Contribution Evaluation in Federated/Machine Learning}
Ghorbani \etal \cite{ghorbani2019data}  and Jia \etal  \cite{pmlr-v89-jia19a} have established that Shapley value  (See Sect.  IV.A) satisfies several important properties (balance, symmetry, zero elements, and additivity) to measure the value of data points in a dataset \cite{shapley1953value}. The main focus of their works is to reduce the time complexity of computing SV. Song \etal \cite{9006327} and Wang \etal \cite{wang2020principled} demonstrate that SV method in FL setting has comparable performance to the centralized ML setting. However, they all assume that a semi-trusted server will honestly evaluate the contribution of each participant. 

\subsection{Privacy-preserving Federated Learning}
There are two lines of works in the \blue{area to ensure client-level privacy:} 1) Local Differential Privacy (LDP) based approach \cite{wang2019collecting,liu2020fedsel, liu2020flame}: \blue{Adding} calibrated noises to the gradients before sending to the server. However, the accumulated noises make the model not very useful. 
It is still an open problem of achieving a good balance between privacy and utility in LDP-based FL.  2) Cryptography-based approach: Secure Multi-Party Computation (MPC) 
and Trusted execution environments (TEE) \cite{cryptoeprint:2016:086}. \blue{LDP adds noise to the final model and possibly degenerates a data owner's contribution. The cryptography-based approach can perfectly protect the intermediate data during model building, but it comes with an expensive computation.} 
MPC is known for several decades that any function can be securely computed; however, it is usually not applicable when the number of parties is large, which is not the case in our setting (See Sect. III). TEE requires extra hardware support and additional engineering steps. Hence, \blue{we adopt MPC in this paper}. Since the native SV method cannot be computed on the masked data, we design an MPC-compatible method to calculate SV. \blue{We recognize the final model privacy is an orthogonal problem to our setting and can be addressed by adding distributed DP noise during MPC computation\cite{7286780}. }

\subsection{Machine Learning on the blockchain}
Chen \etal \cite{8622598} propose a blockchain-based machine learning method using an l-nearest aggregation algorithm to protect the system from potential Byzantine attacks. Nevertheless, they \blue{do not} address the contribution evaluation problem. 
Cai \etal \cite{cai20202cp} introduce a framework to record model updates and contributions on the blockchain. However, the evaluation process is done off-chain, so it is unverifiable, and the framework does not have any privacy guarantee. Our on-chain evaluation method is fully transparent, verifiable, and protects data owners' privacy.

\section{Problem Formulation}
Our target scenario is cross-silo FL, 
and the dataset is horizontally partitioned. 
Moreover, we assume all data owners participate in all training rounds. Our protocol is agnostic to blockchain implementation. Any blockchain with Smart contract capability (e.g., Ethereum, Hyperledger) is compatible with our protocol. Smart contract is a transaction protocol that runs in the blockchain to execute program logic. Indeed, in our setting, Smart contract builds the FL model and evaluates the contribution.

In our framework, any data owner can have two roles: FL model trainer and blockchain miner. The latter role constructs a P2P network with the blockchain protocol that conceptually replaces the traditional centralized server in FL. The protocol has two parts: 1) The leader selection protocol periodically selects a leader to propose a set of transactions. 2) A verification protocol requires all other miners to re-execute the proposed transactions. If the re-execution results are the same as the proposed, the miners accept them; otherwise, they wait for another leader to propose. The blockchain protocol guarantees the transactions' truthfulness as long as the majority of miners are honest. 
\subsection{\blue{Threat/Trust} Model}
In the FL model-building phases, we assume the data owners are semi-honest. They submit honest local weight updates to the blockchain, but they are curious about others' private information since they most likely are competitors or have competing interests. The study \cite{NEURIPS2019_60a6c400} has shown that a curious user can re-build the others' local model if he/she has access to others' local updates, and he/she will be able to retrieve private information from the individual local model. 
In the contribution evaluation phases, the selected data owner (a.k.a leader) may be fraudulent, and he/she will try to maximize his/her contribution by proposing incorrect evaluation results. However, when the majority of miners are honest,  only truthful results are accepted by the blockchain.  Our framework can effectively replace the semi-trusted server in the traditional FL setting.

\section{Our Approach}
\subsection{Preliminaries}
\subsubsection{Secure aggregation} We \blue{adopt} Google's work on secure aggregation \cite{10.1145/3133956.3133982}. Readers may refer to the original work for the security and the privacy guarantee. The secure aggregation is based on discrete logarithm cryptography. Here we use a simple example to illustrate the procedure. 
\begin{itemize}[noitemsep,topsep=0pt]
  \item User $A,B,C$ generate the private key $a,b,c$ and broadcast their public key $g^a, g^b, g^c$ to the blockchain.
  \item User $A,B,C$ computes the Diffie–Hellman key $g^{ab}, g^{bc}, g^{ac}$ based on others' public key.
  \item At each round $r$, each user uses a pseudorandom number generator $PRNG(\cdot)$ to generate a paired masks based on Diffie–Hellman key and submits masked local model updates $w_i$ to the blockchain. For user A: $PRNG(g^{ab}, r) \rightarrow m_{ab}^r$,  $PRNG(g^{ac}, r) \rightarrow m_{ac}^r$. It submits $w_A + m^r_{ab} - m^r_{ac}$.
  \item At each round $r$, the blockchain aggregates all masked updates and the masks are cancelled out.
  $w_A + m^r_{ab} - m^r_{ac} + w_B + m^r_{bc} - m^r_{ab} + w_C + m^r_{ac} - m^r_{bc} = w_A + w_B + w_C$
\end{itemize}

Using secure aggregation, the local updates on the blockchain are masked. The curious user cannot learn others' models. 

\subsubsection{Shapley value}
Shapley value, named in honor of Lloyd Shapley, is a solution concept from cooperative game theory \cite{shapley1953value}. 
The SV of user $i$ with respect to the utility function $u(\cdot)$, denoting as $v_{i}$,  is defined as the average of marginal contribution of $i$ to coalition $S$ over all $S \subseteq I \setminus {i}$. The native method to calculate SV is: 
\begin{equation}
v_i = \frac{1}{n} \sum_{S \subseteq I \setminus \{i\}} \frac{1}{{{n-1} \choose |S|}}[u(S \cup \{i\}) - u(S)] \label{eq1}
\end{equation}
where $n$ is the total number of users.

\setlength{\textfloatsep}{0pt}
\begin{algorithm}
\SetAlgoNoLine
\SetAlgoNoEnd
\KwIn{users $i \in I$, masked user local weights $[w_i]$, random seed $e$, round number $r$,  utility function $u(\cdot)$, number of groups $m$.}
\KwOut{users contributions $[v_i^r]$, global model $W_G$}
\tcp{e.g., $I$ is $[A,B,C,D,E,F,G,H,I]$}
\tcp{e.g., $\pi = A,E,H,B,F,I,C,G,D$}
$\pi \leftarrow permutation(e,r, I)$\;
\tcp{$grouping(\cdot)$ assigns users into groups based $\pi$ and $m$.}
\tcp{$G_{j}$ denotes the $j^{th}$ group (e.g., $m=3$, $G_1$ is $[A, E, H]$).}
$G = \{G_j | j \in [1,m]\} \leftarrow grouping(\pi, m)$\;
\lFor{$j = 1$ \textbf{to} $m$}{
$W_j = \frac{1}{||G_j||}\sum_{i \in G_j}{w_i}$
}
\tcp{$\mathbb{P}(\cdot)$ is a powerset operation.}
\lFor{$ S \subseteq \mathbb{P}(G) $}{
    $W_S = \frac{1}{||S||}\sum_{j \in S}{W_j}$
}
\For{$j = 1$ \textbf{to} $m$}{
\tcp{$V_j$ denotes the SV of $j^{th}$ group.}
$V_j \leftarrow \sum_{S \subseteq G \setminus \{j\}} \frac{1}{m {{m-1} \choose |S|} }[u(W_{S \cup \{j\}}) - u(W_S)]$\;
    \lFor{$i \in G_j$}{
    $v_i^r \leftarrow \frac{1}{||G_j||}V_j$
    }
}
 \caption{\textbf{Group SV}: Contribution Evaluation}
 \label{algo1}
\end{algorithm}

\subsection{Procedure}
In secure aggregation, the masking technique randomize each user's updates to protect users' privacy during the training phase. 
However, this method prevents us from computing SV using native SV calculation (i.e., Eq. \ref{eq1}) since we are not able to calculate the correct utility of excluding a single user (e.g., $u(S \cup \{i\}) - u(S)$).
We propose a group-based SV method, GroupSV. 
First, we partition users $I$ into $m$ groups based on a permutation sample at round $r$ with random seed $e$ (line 1-2), then \blue{use secure aggregation to jointly train a group model for each group} (line 3). From those group models, we build coalition models $W_S$ \blue{of different subsets of groups $G$} using plain aggregation (line 4). Last, we compute each group's SV and assign it to its members (line 5-7). \blue{In other words, each data owner's SV is approximated by its group's SV.}
In following section, we describe the procedure of our protocol.

\begin{itemize}[noitemsep,topsep=0pt]
    \item At the off-chain setup stage, users reach a consensus on FL parameters (e.g., FL algorithm), secure aggregation parameters (e.g., generator $g$) , and contribution evaluation parameters (e.g., permutation seed $e$, group size $m$, utility function $u$) and submit them to the blockchain. 
    \item At each round $r$, users train their models and send masked updates to the blockchain. Then, users download the new global model \blue{(an aggregation of all the group model)} and start the next round until reaching the maximum number of rounds.
    \item In the end, all users share an aggregated global model $W_G$, and the total SV for a user is $v_i = \sum_{r=0}^{R} v_i^r$, where $R$ is the total number of the rounds.  
\end{itemize}

Group SV is configurable with different number of groups, $m$. When $m$ is the maximum, $m = n$ (where $n=|I|$), all users are assigned to a single group, and their SVs are evaluated independently based on their per round local model, so SVs have the highest resolution. However, the model parameters are revealed. In general, given the number of groups $m$, the average model parameters for each group of size $n/m$ is revealed, in some sense similar to $(n/m)$-anonymity.  Hence, the larger the $m$, the less private.  When $m$ decreases, one group has multiple users, and their SVs are computed based on the averaged model, so the resolution decreases. When computing SV using the native method, we need to train additional $2^n - 1$ coalition models. For our group model aggregation, we adopt a similar approach from \cite{9006327}. The coalition models are aggregated from users' local model updates in a FL fashion. 

\section{Experiments}
\subsection{Setup}
\subsubsection{\textbf{Dataset}}
We experiment on the \textit{Optical Recognition of Handwritten Digits} 
dataset \cite{Dua:2019}, which is normalized bitmaps of handwritten digits.  It contains 5620 instances, 64 attributes, and 10 classes. We randomly split the dataset into a training dataset and a testing dataset with a ratio of 8:2 and randomly split the training dataset into 9 subsets to simulate 9 data owners, $D = \{d_i\}_{i=0}^8$. 
To simulate different data quality of each data owner, we add Gaussian noise with an increasing sigma, $d_i = d_i + \mathcal{N}(0, \sigma*i)$. 
As a result, $d_0$ has the best data quality, $d_1$ has worse data quality, and so on.
When $\sigma= 0$, all data owners have similar data quality; the higher $\sigma$ leads to more diverse data quality.
\subsubsection{\textbf{Environment}} 
We conduct experiments on Ubuntu 18.04.4 with Intel(R) i7-6700K CPU @ 4.00GHz and 64GB main memory. The codes are implemented on Python 3.7 with NumPy. We use logistic regression with gradient descent in local train epoch and FedAvg \cite{mcmahan2017communication} in global train epoch. The contribution evaluation phase occurs in a simulated blockchain. 
\subsection{Results}
We next analyze the three experimental results.
\vspace*{-10pt}
\begin{figure}[htbp]
\centerline{\includegraphics[width=6.6cm]{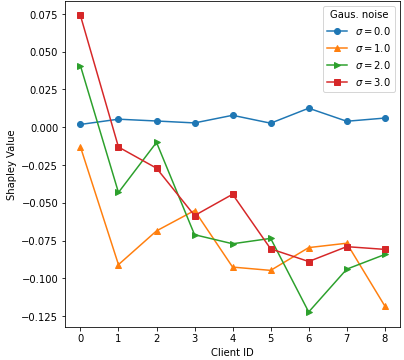}}
\vspace*{-10pt}
\caption{Ground Truth SV distribution over users w.r.t. different $\sigma$.}
\label{native}
\end{figure}
\vspace*{-15pt}

\begin{figure}[htbp]
\centerline{\includegraphics[width=6.4cm]{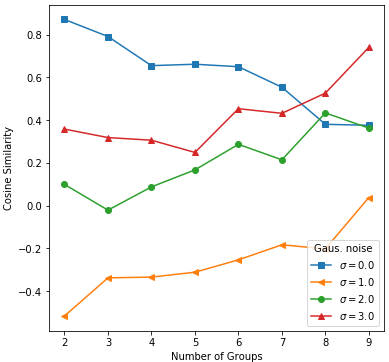}}
\vspace*{-10pt}
\caption{The approximation accuracy between Group SV method and Native SV method with different Gaussian noise.}
\label{group}
\end{figure}

\vspace*{-15pt}

\begin{table}[htbp]
\caption{Time comparison between GroupSV and NativeSV}
\begin{tabular}{c|c|c|c|c|c|c|c|c|c|}
\cline{2-10}
 & \multicolumn{8}{c|}{GroupSV} & NativeSV \\ \cline{2-10} 
\# of Groups & 2 & 3 & 4 & 5 & 6 & 7 & 8 & 9 & 9 \\ \cline{2-10} 
Time/s & 2 & 3 & 4 & 7 & 11 & 20 & 39 & 77 & 316 \\ \cline{2-10} 
\end{tabular}
\label{tab1}
\end{table}
\subsubsection{\textbf{Ground Truth}}
First, we build $2^n$ models based on the data coalitions, $\{M_S | S \subseteq \mathbb{P}(I) \}$, then establish the ground truth SV using the native SV method (Eq. \ref{eq1}). 
\blue{We emphasize that native SV cannot be computed with privacy protection on the blockchain.}

Fig.\ref{native} shows users' SVs with respect to different Gaussian noise.
When there is no noise added ($\sigma = 0$), all users have a close to zero SV. We split the dataset uniformly at random, so the subsets have similar distributions, and the marginal contributions towards the final model are negligible. As a result, they contribute almost equally to the final model. When $\sigma \neq 0$, the good (less noisy) quality dataset has a higher SV than the bad quality dataset. 
As we expected, SV can distinguish different data owners' contribution according to their quality.

\subsubsection{\textbf{Accuracy Comparison}}
We use cosine similarity, i.e., $similarity = \cos{\theta} = \frac{\vec{u} \cdot \vec{v}}{|\vec{u}||\vec{v}|}$ , to measure the distance between the Shapley values calculated by our group SV and the ground truth SV.
Fig. \ref{group} shows the change of cosine similarity with respect to the number of groups. 
We can see that, when $\sigma= 0$ (i.e., all data owners have similar data quality), the similarity is decreasing along with the number of groups. 
The reason is that, when $\sigma= 0$, the ground truth SV value distribution is close to uniform over the participants.
Note that, our group-based SV calculation allocates contribution uniformly to each participant within the group.
Hence, when the group size is large (i.e., the number of groups is small), most users have the similar SV, so the cosine similarity in this case is high. 
On the other hand, we observe that, when $\sigma \neq 0$, the similarity increases with the number of groups. 
It is because, when the number of groups is large, our group based method is closer to the native one. 
For the same reason, the cosine similarity increases with respect to increasing $\sigma$ (i.e., more diverse data quality).

\subsubsection{\textbf{Runtime Comparison}}
Tab. \ref{tab1} shows the time performance of our group SV and the native SV method. Since coalition models are aggregated directly from users' local updates, the number of models required to train to compute SV reduces from $2^n$ to $n$. Consequently, we see an order of magnitude improvement in time in our group SV method. 

\section{Conclusion \& Future Work}
This paper addresses transparent contribution evaluation challenges for different data owners in a cross-silo horizontal FL setting. We propose a blockchain-based method to measure data owners' SV-based contributions with configurable resolution without sacrificing their privacy. 
Our experiments on the handwriting digits dataset demonstrate that our method can effectively evaluate contributions.

Although we proposed core algorithms and demonstrated their effectiveness, several important future works are needed to complete this study.
First, we will investigate the proposed method's applicability to the existing blockchain platforms (such as Ethereum or Hyperledger Fabric).
We will pinpoint the potential bottlenecks (such as transaction throughput) of implementing secure federated learning with the blockchain.
Second, we will study the effects of adversarial participants on the Shapley value calculation since the proposed group-based SV method may be influenced by the number of groups and the participants' adversarial behavior.
Third, we will thoroughly examine the trade-offs between privacy, transparency, and security in the proposed system. 

\section*{acknowledgment}
This work is supported by National Science Foundation (NSF) CNS-1952192, National Institutes of Health (NIH) R01GM118609, JSPS KAKENHI Grant No. 19K20269, CCF-Tencent Open Fund WeBank Special Fund.

\bibliographystyle{ieeetr}
\bibliography{main.bib}{}

\end{document}